\documentclass[conference]{IEEEtran}
\usepackage{times}
\usepackage[numbers]{natbib}
\usepackage{multicol}
\usepackage[bookmarks=true]{hyperref}
\usepackage{booktabs}      
\usepackage{threeparttable} 
\usepackage{multirow} 
\usepackage{amssymb,amsfonts,bm}
\usepackage{soul}
\usepackage{graphicx}
\usepackage{textcomp}
\usepackage{amsmath,tikz}
\usetikzlibrary{tikzmark}
\usepackage{color}
\usepackage{url}
\usepackage{mathtools}
\usepackage{rotating}
\usepackage{blkarray}
\usepackage{cleveref}
\usepackage{physics}
\usepackage{tabularx}
\usepackage{float}
\usetikzlibrary{arrows}
\usepackage{placeins}
\let\proof\relax
\let\endproof\relax
\usepackage{amsthm}

\usepackage{caption}
\usepackage{subcaption}
\usepackage{etoolbox}
\let\classAND\AND
\let\AND\relax
\usepackage{algorithm}
\usepackage{algorithmicx,algpseudocode}

\let\AND\classAND
\AtBeginEnvironment{algorithmic}{\let\AND\algoAND}

\graphicspath{{eps/}{png/}}
\DeclareSymbolFont{symbolsC}{U}{pxsyc}{m}{n}
\DeclareMathSymbol{\coloneqq}{\mathrel}{symbolsC}{"42}

\newcommand{\vertiii}[1]{{\left\vert\kern-0.25ex\left\vert\kern-0.25ex\left\vert
		#1 
		\right\vert\kern-0.25ex\right\vert\kern-0.25ex\right\vert}}
\usepackage{multirow}
\usepackage{makecell}
\usepackage{adjustbox}
\usepackage{float}

\DeclareMathOperator{\softplus}{SoftPlus}

\newcommand*{\vecbf}[1]{\mathbf{#1}}
\newcommand*{\matbf}[1]{\mathbf{#1}}
\newcommand*{\gbf}[1]{\bm{#1}}
\newcommand*{\myset}[1]{\mathcal{#1}}

\usepackage{pifont}

\usepackage[scaled=0.92]{helvet}
\newcommand*{\Push}{\textsf{Push-slide-settle}}
\newcommand*{\Fall}{\textsf{Fall-and-rebound}}

\newcommand*{\Mass}{\matbf{M}}

\newcommand*{\pos}{\vecbf{p}}
\newcommand*{\quat}{\vecbf{q}}
\newcommand*{\vel}{\vecbf{v}}
\newcommand*{\samplingTime}{h}
\newcommand*{\jac}{\matbf{J}}
\newcommand*{\dist}{\phi}
\newcommand*{\fric}{\mu}
\newcommand*{\Dist}{\gbf{\phi}}
\newcommand*{\gainK}{\matbf{K}}

\newcommand*{\force}{\gbf{\tau}}
\newcommand*{\forceb}{\vecbf{b}}
\newcommand*{\contIm}{\gbf{\lambda}}
\newcommand*{\image}{\matbf{I}}
\newcommand*{\gmap}{\matbf{G}}
\newcommand*{\action}{\matbf{a}}
\newcommand*{\damp}{\matbf{D}}
\newcommand{\dvec}{\matbf{d}}
\newcommand{\nvec}{\matbf{n}}

\newcommand*{\allpara}{\Theta}
\newcommand*{\camera}{\matbf{T}_{cw}}
\newcommand*{\stat}{\vecbf{x}}
\newcommand*{\ctrl}{\vecbf{a}}
\newcommand*{\dyn}{\vecbf{f}}
\newcommand{\R}{\mathbb{R}}

\newcommand*{\loft}{\mathrm{Loft}}

\newcommand*{\gauss}{\mathcal{G}}
\newcommand*{\para}{\gbf{\theta}}

\newcommand*{\timeSet}{\myset{T}}
\newcommand*{\loss}{\myset{L}}

\newcommand*{\cen}{\vecbf{c}}

\newcommand*{\friction}{\mu}

\newcommand*{\transf}{\mathrm{TF}}
\newcommand*{\queryP}{\vecbf{p}}

\begin{document}

\title{ContactGaussian-WM: Learning Physics-Grounded World Model from Videos}
\author{
    Meizhong~Wang, Wanxin~Jin, Kun~Cao, Lihua~Xie, Yiguang~Hong
	}
   
\makeatletter
\g@addto@macro\@maketitle{
\setcounter{figure}{0}
  \vspace{-10pt}
  \begin{figure}[H]
  \setlength{\linewidth}{\textwidth}
  \setlength{\hsize}{\textwidth}
  \centering
  \includegraphics[width=\textwidth]{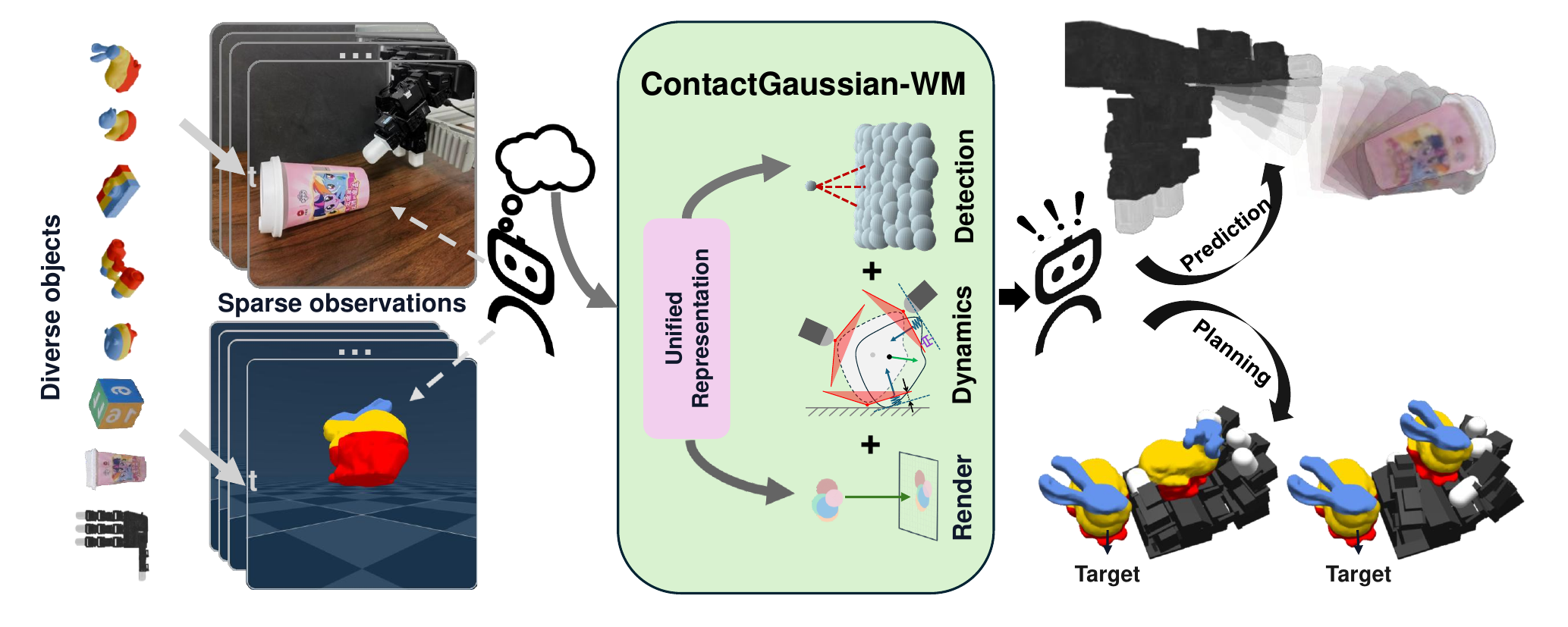} 
  \caption{\small{ContactGaussian-WM integrates a differentiable collision detector, explicit contact dynamics, and a 3DGS renderer. It takes sparse videos of diverse objects as input, adopts a unified scene representation, and learns a physics-grounded world model that supports downstream tasks such as long-horizon prediction and real-time MPC for in-hand manipulation.}}
   \label{fig:teaser}
  \end{figure}
  \vspace{-15pt}
}

\maketitle

{\renewcommand{\thefootnote}{}\footnotetext{M. Wang, K. Cao (corresponding author), and Yiguang Hong are with the Department of Control Science and Engineering, College of Electronics and Information Engineering, Tongji University, Shanghai 201804, China,  the Shanghai Institute of Intelligent Science and Technology, National Key Laboratory of Autonomous Intelligent Unmanned Systems, and Frontiers Science Center for Intelligent Autonomous Systems, Ministry of Education, Beijing 100816, China{\tt \footnotesize \{2531784,caokun,yghong\}@tongji.edu.cn}\\
    W~Jin is with the School for Engineering of Matter, Transport, and Energy, Arizona State University. {\tt \footnotesize wanxinjin@gmail.com} \\
L.~Xie is with School of Electrical and Electronic Engineering, Nanyang Technological University, 50 Nanyang Avenue, Singapore 639798 {\tt \footnotesize elhxie@ntu.edu.sg}\\
  }}

\begin{abstract}
 Developing world models that understand complex physical interactions is essential for advancing robotic planning and simulation. 
 However, existing methods often struggle to accurately model the environment under conditions of data scarcity and complex contact-rich dynamic motion. 
 To address these challenges, we propose ContactGaussian-WM, a differentiable physics-grounded rigid-body world model capable of learning intricate physical laws directly from sparse and contact-rich video sequences. 
 Our framework consists of two core components: (1) a unified Gaussian representation for both visual appearance and collision geometry, and (2) an end-to-end differentiable learning framework that differentiates through a closed-form physics engine to infer physical properties from sparse visual observations.
 Extensive simulations and real-world evaluations demonstrate that ContactGaussian-WM outperforms state-of-the-art methods in learning complex scenarios, exhibiting robust generalization capabilities. 
 Furthermore, we showcase the practical utility of our framework in downstream applications, including data synthesis and real-time MPC. Project page: \url{https://contactgaussian-wm.github.io/}.
\end{abstract}

\IEEEpeerreviewmaketitle

\section{Introduction} \label{sec:intro}
World models \cite{https://doi.org/10.5281/zenodo.1207631} have emerged as the cognitive engine for autonomous robotics, serving as internal simulators that enable agents to reason about the physical environment. 
By synthesizing high-dimensional visual observations into predictive representations, these models allow robots to anticipate complex physical behaviors—such as motion, interaction, and deformation—within their `mind’s eye', facilitating efficient planning without costly real-world trial and error. 
However, learning such world models directly from sparse and contact-rich video sequences remains a formidable challenge.

Current mainstream paradigms for world models, predominantly purely data-driven approaches like video generation~\cite{brooks2024video,hu2023gaia,bar2024lumiere,wang2024driving} and latent dynamics models~\cite{ha2018world,hafner2024masteringdiversedomainsworld,wu2023daydreamer,bardes2024revisiting,hansen2023td}, face significant hurdles in practice. 
These methods are notoriously data-hungry and often fail to generalize in scenarios characterized by data scarcity or few-shot demonstrations. 
More critically, without explicit physical grounding, these models prioritize visual plausibility over physical rigor, leading to hallucinations such as object interpenetration or spontaneous disappearance. 
This absence of inherent constraints further results in a fundamental inability to maintain long-horizon consistency, critical to downstream tasks such as planning. 

To overcome these limitations and mitigate the dependency on massive datasets, the research community is shifting toward physics-grounded world models that internalize the invariant laws of nature to ensure long-horizon reliability. 
Leveraging high-performance simulators (e.g., MuJoCo \cite{todorov2012mujoco}, Bullet \cite{coumans2016pybullet}) allows for the rigorous simulation of physical laws, which, when coupled with renderers, can be extended to image-level world models. 
However, these frameworks are typically non-differentiable; consequently, to align with the real world, some approaches \cite{yang2025twintrack,ramos2019bayessim,memmel2024asid,fan2025twinalignervisualdynamicalignmentempowers} must employ gradient-free optimization for parameter identification. 
This paradigm, however, is heavily reliant on data quality while suffering from prohibitive search spaces and convergence difficulties. 

Furthermore, increasing research efforts are dedicated to building differentiable world models, aiming to enable stable and efficient system identification via gradient descent \cite{li2025pinwmlearningphysicsinformedworld,fan2025twinalignervisualdynamicalignmentempowers,moran2025splattingphysicalscenesendtoend,anonymous2025drex,jatavallabhula2021gradsim}. 
However, achieving stable end-to-end learning remains challenging.
Methods like \cite{li2025pinwmlearningphysicsinformedworld} utilize integrated simulators for collision detection, rendering the collision process non-differentiable. 
In contact-rich interactions where object trajectories are hypersensitive to the precise contact manifold, this `gradient detachment' prevents the model from effectively correcting geometric errors. 
Consequently, the optimization landscape becomes notoriously rugged and ill-posed, making it difficult to recover precise dynamic parameters from sparse and contact-rich video sequences.
While approaches such as \cite{moran2025splattingphysicalscenesendtoend,anonymous2025drex,jatavallabhula2021gradsim} achieve theoretical end-to-end differentiability, they essentially rely on numerical approximation and lack analytical gradients, and hence are unsuitable for the learning and application of high-quality world models.
Moreover, due to the structural decoupling of perception and dynamics, these methods are often compelled to maintain two distinct representations: one for visual rendering (3DGS~\cite{kerbl20233d} or Nerf \cite{mildenhall2021nerf}) and another for physical simulation (e.g., explicit meshes or particles). 
In practice, enforcing precise geometric alignment between these dual representations is non-trivial, causing the back-propagated gradients from visual reconstruction errors to be indirect and spatially misaligned with the underlying physical parameters.

In this work, we introduce a novel differentiable framework called ContactGaussian-WM for learning a physics-grounded world model from a small number of interactive videos.
This framework enables visually realistic and physically plausible predictions and bridges the sim-to-real gap in robot planning tasks. 
Specifically, we unify visual and geometric representations, where a decoupled optimization approach is used to train 3DGS~\cite{kerbl20233d} to initialize the scene and leads to a representation that balances visual and geometric aspects. 
To ensure the differentiability of collision detection, we force the 3DGS~\cite{kerbl20233d} to be isotropic spheres. 
To adapt to contact-rich scenarios, we select a lightweight yet capable contact dynamics model \cite{jin2024complementarity} that is effective enough to handle different contact modes (skip, separation, stick) \cite{xiepalm,xie2026touchcontacthierarchicalrlmpc}. 
This model, combined with a 3DGS differentiable renderer, is used to construct ContactGaussian-WM, enabling differentiable learning, prediction, and planning control of the world model.

Extensive experiments, in both simulated and real-world environments, demonstrate that our proposed framework is capable of learning a physics-grounded world model from sparse observations in complex contact-rich scenarios. 
The resulting model facilitates long-horizon, accurate, and physically consistent predictions, effectively bridging the sim-to-real gap. 
Furthermore, we showcase its capability for real-time planning when integrated with Model Predictive Control (MPC). 
Our main contributions are summarized as follows:
\begin{itemize}
    \item We propose ContactGaussian-WM, an end-to-end differentiable framework with a unified geometric and visual representation of the world, to achieve efficient learning of world models from sparse videos in contact-rich scenarios; and 
    \item We conduct extensive simulation and real-world experiments, validating our method's generalizability across diverse complex interaction scenarios and demonstrating that it outperforms existing  methods\cite{li2025pinwmlearningphysicsinformedworld,hafner2024masteringdiversedomainsworld}.
\end{itemize}

\section{Related Works}
\subsection{World model}
\subsubsection{Data-driven world model}
Recent generative AI has revitalized pixel-based approaches, where foundation models like ~\cite{bruce2024genie,brooks2024video,hu2023gaia,yang2023unisim,bar2024lumiere,wang2024driving} serve as generalist video simulators. 
To improve control efficiency, latent-state-based methods~\cite{hafner2019learning,ha2018world,hafner2024masteringdiversedomainsworld,wu2023daydreamer,bardes2024revisiting,hansen2023td} compress observations into compact states. 
However, both paradigms treat dynamics as black-box transitions of probability distributions. 
Lacking explicit physical constraints, they suffer from physical hallucinations, e.g., object morphing or vanishing, since they essentially mimic data statistics rather than enforcing causal physical laws.
\subsubsection{Physics-grounded world models}
To improve physical plausibility, some works incorporate structural priors using graph neural networks on particle representations \cite{li2018learning, wu2023slotformer}. While these offer better generalization, they remain "black-box" and lack strict guarantees for non-penetration or friction. Alternatively, some methods integrate explicit physics engines with renderers \cite{yang2025twintrack, ramos2019bayessim}, but rely on gradient-free optimization which scales poorly with search space.

Consequently, recent research explores differentiable world models. PIN-WM \cite{li2025pinwmlearningphysicsinformedworld} achieves local differentiability via implicit differentiation of an LCP solver, while others \cite{moran2025splattingphysicalscenesendtoend, anonymous2025drex, jatavallabhula2021gradsim} use numerical approximations. However, these often suffer from gradient discontinuities at contact boundaries and visibility changes, leading to unstable training. Furthermore, non-differentiable collision detection in models like PIN-WM prevents geometry optimization, making them unsuitable for contact-rich scenarios where geometric errors cannot be corrected. In contrast, ContactGaussian-WM achieves true end-to-end differentiability, enabling the simultaneous optimization of physical parameters and collision geometry.

\subsection{Rigid Body Dynamics Simulation}
\subsubsection{Collision detection}

Mainstream physics engines (e.g., Bullet, MuJoCo, DART, and Drake) predominantly rely on mesh-based discrete representations and algorithms like GJK~\cite{gilbert2002fast} and EPA~\cite{van2003collision} to compute separation distances. While precise for forward simulation, these methods are non-differentiable and ill-suited for gradient-based optimization. 

To address this, recent frameworks have shifted toward continuous geometric representations. Methods like gradSim~\cite{jatavallabhula2021gradsim}, D-REX~\cite{anonymous2025drex}, and ContactSDF~\cite{yang2025contactsdfsigneddistancefunctions} utilize volumetric fields (SDFs or density grids) to support gradient propagation through field penetration. Similarly, recent work~\cite{moran2025splattingphysicalscenesendtoend} employs 3DGS~\cite{kerbl20233d} to model geometry for physical interactions. However, these methods often rely on spatial and temporal discretization and approximation during contact resolution. These numerical approximations introduce gradient discontinuities or inaccuracy that limit detection accuracy and fall short of the fidelity provided by exact analytical solutions.

\begin{figure*}[!htbp]
    \centering
    \includegraphics[width=1.0\textwidth]{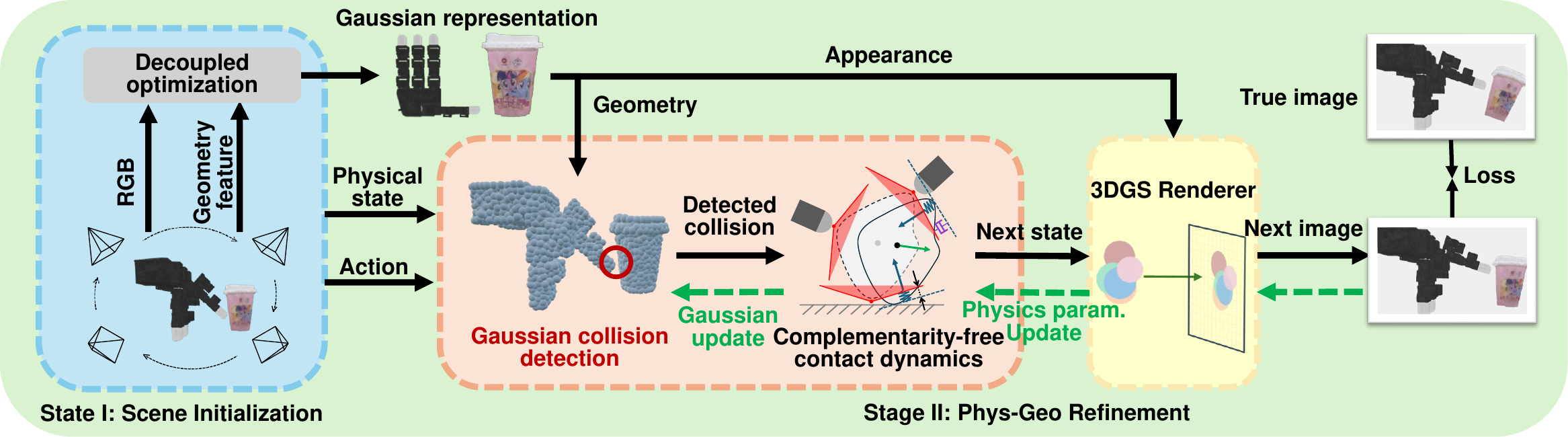} 
        \caption{Overview of ContactGaussian-WM. We first initialize a scene with a unified spherical Gaussian representation, then jointly refine physics and geometry: given the current physical state and action, the differentiable collision detector uses Gaussian geometry to compute contact points, which are fed into the complementarity-free contact dynamics model to compute the next physical state, and the 3DGS renderer generates the next image. The pipeline is fully differentiable for end-to-end learning.}
    \label{fig:overview of ContactGaussian-WM}
\end{figure*}

\subsubsection{Contact Models}
Traditional rigid-body contact models are often formulated as Nonlinear Complementarity Problems (NCP) \cite{stewardtrinklePolyhedron, stewart2000implicit, stewart1996implicit}, which is computationally inefficient for high dimensions.
One common approach to simplify its computation is to approximate Coulomb friction cones by polyhedral cones \cite{stewart1996implicit, todorov2012mujoco}, allowing the formulation to be transformed into Linear Complementarity Problems (LCP) solvable by efficient solvers. 
Another approach is to relax the NCP formulation \cite{anitescu2006optimization} by constraining the contact velocity within a dual friction cone, transforming the problem into Convex Complementarity Problems (CCP).
This practice offers significant computational benefits, such as fast convergence and theoretical solution guarantees \cite{anitescu2010iterative, tasora2011matrix}. 
Furthermore, the aforementioned CCP can be formulated as the Karush-Kuhn-Tucker (KKT) optimality conditions for a second-order cone-constrained convex optimization problem \cite{anitescu2006optimization,pang2023global, pang2021convex, todorov2012mujoco}. 
To endow the learnability for the model, several approaches \cite{geilinger2020add, howell2022dojo, todorov2010implicit, yamane2006stable, liu2024quasisim} have proposed differentiable approximations as alternatives to complementarity-based models. 
These methods achieve differentiability through various operations, such as introducing penalty functions \cite{geilinger2020add, yamane2006stable, liu2024quasisim}, relaxing complementarity constraints \cite{howell2022dojo}, or employing implicit complementarity formulations \cite{todorov2010implicit}. 
A shared feature of these methods is that they ultimately require iteratively solving a residual equation for contact constraint resolution, with differentiability typically being obtained via the Implicit Function Theorem \cite{rudin1964principles}. 
Consequently, some researchers in \cite{tassa2012synthesis, chatzinikolaidis2020contact, yamane2006stable, liu2024quasisim, jin2024complementarity} have explored closed-form contact dynamics models. 
While some works \cite{tassa2012synthesis, chatzinikolaidis2020contact, yamane2006stable, liu2024quasisim} seek a closed-form solution by decoupling contact normal forces from Coulomb friction, this often introduces additional hyperparameters and potential non-physical artifacts. 
In contrast, the approach in \cite{jin2024complementarity} utilizes an approximation within the dual friction cone. 
This allows for a unified treatment of contact normal and frictional forces within a single term, ensuring flexible capture of different contact modes and automatic satisfaction of Coulomb’s friction law.

In this work, our rigid body simulator implements differentiable collision detection and integrates a closed-form contact model \cite{jin2024complementarity}, creating a more efficient differentiable simulation and providing stable analytical gradients.

\section{Methods}
\subsection{Overview of our framework}
Figure~\ref{fig:overview of ContactGaussian-WM} outlines the ContactGaussian-WM framework, which mainly comprises two stages: scene initialization and phys-geo refinement. 
Scene initialization uses decoupled optimization to obtain a unified spherical Gaussian representation, balancing visual and geometric aspects. 
Phys-geo refinement is achieved through the ContactGaussian-WM world model, which mainly consists of a differentiable collision detector, an explicit contact dynamics model, and a 3DGS renderer. 
All those components are discussed in detail as follows.

\subsection{Problem formulation}
\label{sec:prob}
Given an RGB video of a contact-rich scene, our goal is to build a world model that predicts the future images of the scene and captures the underlying physical laws.
Denote the image of the scene at time step $t$ as $\image_{t}$. 
Our goal is to minimize the difference between the image $\hat\image_{t}$ predicted by our world model and the actual image $\image_{t}$.
In our ContactGaussian-WM, we represent the contact-rich scene by Gaussians, denoted $\gauss:= \{\gauss_{\text{geo}}, \gauss_{\text{vis}}\} = \{\{\mathbf{c}, s\},\{\alpha, f\}\}$, where $\gauss_{\text{geo}}$ (resp. $\gauss_{\text{vis}}$) encodes the geometry (resp. appearance) of all objects in the scene with the spatial center and the isotropic scale (opacity $\alpha$ and color features $f$) of the Gaussian primitives.
ContactGaussian-WM predicts future images of the scene by transforming the Gaussian $\gauss$ by the physical state $\hat\stat_{t}$, and rendering them. We denote this process using a function $\R$.
The evolution of the physical state $\hat\stat_{t}$ over time is realized using our differentiable simulator (Section \ref{sec:phys}). 
Therefore, the optimization problem is formulated as:
\begin{equation}
\label{eq:prob}
\begin{aligned}
\min_{\gauss, \para} & \quad \mathcal{L}(\gauss, \para) = \sum_{t} \left( \mathcal{L}(\hat{\image}_{t}, \image_{t}) \right) \\
\mathrm{s.t.} & \quad \hat{\stat}_{t+1} = \dyn(\hat{\stat}_{t}, \action_{t}, \para, \gauss_{\text{geo}}), \quad \hat{\image}_{t} = \R(\hat{\stat}_{t}, \gauss),
\end{aligned}
\end{equation}
where $\dyn$ is our differentiable simulator with control input $\action_{t}$ and being parameterized by physical properties $\para$ (detailed later) and Gaussian geometry attributes $\gauss_\text{geo}$. 
This cost function is designed as $\loss = \loss_{\loft} + \loss_{\mathrm{L1}}$, which includes Loft feature loss \cite{sun2021loftrdetectorfreelocalfeature} and L1 loss to help us better capture differences at both the feature and pixel levels.

Jointly learning Gaussian parameters $\gauss$ and physical parameters $\para$ from a single video segment by Eq.\eqref{eq:prob} is very challenging due to the high nonlinearity and ambiguity.
To address this, we divide the entire learning framework into two stages. 
The first stage focuses on optimizing the Gaussians $\gauss$, while the second stage is to optimize the physical parameters $\para$, and simultaneously fine-tunes the geometry attributes $\gauss_{\text{geo}}$.

\subsection{Stage I: Scene Initialization with Sphere Geometry Gaussian Splatting (SG-GS)}
We first obtain multi-view observations $\{\image_{i}\}$ of the scene from single or multiple calibrated cameras. 
Based on that, we use SAM2 \cite{ravi2024sam2segmentimages} to extract geometry features $\gmap_{i}$.
Then, we propose an \textbf{SG-GS} method to obtain a Gaussian representation of the scene from the observation, which has two key designs:

\textbf{Isotropic spherical primitives.} 
Unlike traditional 3DGS ~\cite{kerbl20233d}, we enforce isotropic scaling and frozen rotation updates of Gaussians to ensure that each Gaussian maintains a spherical shape, which helps us achieve closed-form differentiable collision detection in Stage II.

\textbf{Decoupled optimization.}
To reconcile the conflict between high-frequency visual details and stable physical geometry, we adopt a \textbf{decoupled optimization} strategy. 
First, we focus exclusively on learning the geometric attributes $\gauss_{\text{geo}}$ while keep the visual appearance attributes $\gauss_{\text{vis}}$ frozen. 
This compels the primitives to strictly align with the geometrical structure of the object rather than compensating for photometric errors. 
Consequently, the optimization minimizes the reconstruction error between the rendered and observed geometric maps:
\begin{equation}
    \min_{\gauss_{\text{geo}}}  \quad \mathcal{L}_{\text{geo}} := \sum\nolimits_{i} \| \hat{\gmap}_i(\gauss_{\text{geo}}) - \gmap_i \|_{1}
\end{equation}
where $\hat{\gmap_i}$ and $\gmap_i$ denote the rendered and ground-truth geometric feature maps, respectively.

Second, we refine the visual fidelity by optimizing the visual appearance attributes $\gauss_{\mathrm{vis}}$, while freezing the geometric attributes $\gauss_{\mathrm{geo}}$ to preserve the collision proxies. 
Following the standard 3DGS protocol~\cite{kerbl20233d}, the objective combines L1 and D-SSIM losss \cite{baker2023dssimstructuralsimilarityindex}:
\begin{equation}
    \min_{\gauss_{\text{vis}}}  \quad \mathcal{L}_{\text{vis}} :=  \sum\nolimits_{i} ((1 - \lambda) \| \hat{\image}_i - \image_i \|_1 + \lambda \mathcal{L}_{\text{D-SSIM}}(\hat{\image}_i, \image_i))
\end{equation}
where $\lambda$ is a weighting factor, and $\hat{\image}_i$ is the rendered image.

These modifications yield a unified SG-GS representation that ensures both visual accuracy and geometric robustness. 
By differentiably extracting Gaussian geometry parameters $\gauss_{\text{geo}}$ as explicit collision geometries, we transform Gaussians from mere rendering kernels into active participants in physical simulation. 
This ensures strict alignment between the visual appearance and the physical model, realizing a ``what you see is what you get" framework.

 \subsection{Stage II: Phys-Geo Refinement}
\label{sec:phys}
With the optimized Gaussian $\gauss^*$, we next optimize the physical parameters $\para$ and fine-tune geometric attributes $\gauss_\text{geo}$, based on the gradient flow from visual observations.
To fully represent the dynamics in Eq.\eqref{eq:prob} by a 3D rigid-body simulation system, we define the environment state as $\stat_t = \{\quat_t, \vel_t\}$, representing the system's generalized coordinates and velocity of all objects at time $t$.
Furthermore, the physical properties that govern the state transitions are incorporated into the system learnable physical parameters $\para=($$\Mass$, $\fric$, $\gainK$, $\damp)$, where $\Mass$, $\fric$, $\gainK$, $\damp$ denotes the mass-inertia matrix, friction coefficient, stiffness, and damping coefficient, respectively. 
In this work, 
We jointly optimize its intrinsic physical properties and geometrical attributes by back-propagating through the following three core processes: a differentiable collision detection based on spherical Gaussians, an explicit contact dynamics model, and a differentiable 3DGS rendering.

\noindent
\subsubsection{Differentiable collision detection.} Leveraging the current predicted states $\hat{\stat}_{t}$ and initialized  geometric attributes $\gauss_{\text{geo}}^*$  which represent the collision geometry as a union of $n$ spherical primitives. 
For each primitive $i \in \{1, \dots, n\}$, its center $\mathbf{c}_i$ is derived from the Gaussian's position, and its radius is defined as $r_i = 2s$ (twice the scale radius is a reasonably conservative value). 
At this point, we can use a simple formula to represent the shortest distance from the query point $\queryP$ to the Gaussian collision geometry:
\begin{equation*}
    \dist_{\gauss_{\text{geo}}}(\queryP) = \min_{i \in \{1, \dots, n\}} \left( \| \queryP - \cen_i \|_2 - r_i \right).
\end{equation*} 
To ensure differentiability, we first approximate the shortest distances using the LogSumExp (LSE) smoothed minimum:
\begin{equation*}
    \dist_{\text{soft}}(\queryP) \approx -\frac{1}{\beta} \log ( \sum_{i=1}^{n} \exp \left( -\beta (\| \queryP - \cen_i \|_2 - r_i) \right) ).
\end{equation*}
where $\beta$ is the smoothing factor. 
However, applying the above formula to calculate the distance inside an object often reflects the proximity to the nearest individual sphere rather than the object's actual surface, leading to inaccurate penetration estimates. 
To address this, a significant negative penalty is assigned to ensure a robust correction of inter-penetration when the distance is negative. 
To preserve differentiability for end-to-end learning, we design a smoothing transition with the Sigmoid function $\sigma(\cdot)$:
\begin{equation*}
    \dist(\queryP) \approx \sigma(\gamma \cdot \dist_{\mathrm{soft}}(\queryP)) \cdot \dist_{\mathrm{soft}}(\queryP) + \left( 1 - \sigma(\gamma \cdot \dist_{\mathrm{soft}}(\queryP)) \right) \cdot (-\delta),
    \label{eq:final_collision_dist}
\end{equation*}
where $\gamma$ controls how sharp $\sigma$ is, and $\delta$ is the positive constant penalty used for negative distances inside-object cases.
Using the calculated distance $\dist(\queryP)$, the nearest surface point $\queryP_{\mathrm{c}}$ is obtained via gradient-based projection:
\begin{equation*}
    \queryP_{\mathrm{c}}(\queryP) = \queryP - \dist(\queryP) \cdot \nabla_{\queryP} \dist(\queryP) / \| \nabla_{\queryP} \dist(\queryP) \|_2,
    \label{eq:nearest_point}
\end{equation*}
with the normalized gradient $\nabla_{\queryP} \dist(\queryP)$ being the surface normal. 
Based on this collision information, the contact Jacobian can be computed based on rigid body kinematics \cite{spong2020robot}.

\subsubsection{Differentiable complementarity-free contact dynamics.}
We adopt the full-dynamic complementarity-free contact model \cite{jin2024complementarity} for contact-rich scenarios, which is given by:
\begin{equation*}
\begin{aligned}
    \label{eq:full_dyn_eqn}
    &\vel_{t+1} = \overbrace{(\vel_t+h\Mass^{-1}\force(\mathbf{q}_t,\vel_{t},\ctrl_{t}))}^{:=\forceb_{t}}+
    h\Mass^{-1} \tilde\jac^{\top}_{t} \contIm_{t} \\
    &\contIm_{t} := \softplus\Big(
                \underbrace{-\gainK \big(\samplingTime\tilde{\jac}_{t}\forceb_{t} + \tilde{\Dist}_{t}\big)}_{\text{stiffness of dual cone}}
                \underbrace{
                    - \damp \big(\tilde{\jac}_{t}\forceb_{t})}_{\text{damping of dual cone}}\Big)
                ,
\end{aligned}
\end{equation*}
Here, $\Mass$ represents the generalized mass matrix of the rigid body.
The force vector $\force(\mathbf{q}_t,\vel_{t},\ctrl_{t})$ is the sum of non-contact forces in the system, including actuation force, gravity, and Coriolis forces. $\tilde{\Dist}_{t}$ is the signed collision distance of each collision pair. 
$\tilde\jac_{t}(\mu)$ is the Jacobian of the faces of the polyhedral dual cone (of the friction cones) at each collision pair. 
These are computed  from the raw collision information ($\jac_{t}$, ${\Dist}_{t}$) and frictional coefficient $\friction$ (See Appendix for detailed form). 
Instead of iteratively solving the contact constraint, the complementarity-free contact model approximates the contact force $\mathbf{\lambda}$ based on an impedance mechanism defined on the dual cones, with $\gainK$ and $\damp$ representing the learnable stiffness and damping coefficients. This leads to a closed-form solution for the contact forces $\contIm_{t}$ on the dual cones of the Coulomb friction cones.
Moreover, different contact modes, e.g., sticking, sliding, and separation, can be captured by different penetration regimes across facets of the dual cones, and Coulomb friction laws are automatically satisfied.
With the computed $\vel_{t+1}$, the system's next generalized position $\mathbf{q}_{t+1}$ can be obtained by $\mathbf{q}_{t+1}{=}\mathbf{q}_{t} \oplus \samplingTime\vel_{t+1}$,
where $\oplus$ stands for the integration of system pose (position, orientation) with velocities.

\subsubsection{3DGS rendering}
Up to now, we have a differentiable simulator $\dyn$, which helped us successfully construct the state transition function in Eq.\eqref{eq:prob}. 
Using the current predicted state $\hat{\stat}_{t}$ and action $\action_{t}$, we can explicitly obtain the next predicted state $\hat{\stat}_{t+1}$, thus creating the computational graph at the state space.
In order to perform optimization in the image space, we need to map the system states to that space. 
This can be done by transforming all Gaussians based on forward kinematics with the updated state $\hat{\stat}_{t+1}$, which can be written as
\begin{equation}
    \gauss^*_{\transf}(\hat\quat) = \mathbb{FK}(\gauss^*, \hat\quat).
\end{equation}
where $\mathbb{FK}$ is the forward kinematics transformation that applies a rigid body transformation to the properties of all Gaussians in the scene based on the current state $\hat\quat$, obtaining the Gaussians in the current predicted state. 
Then, based on the camera pose $\camera$ and the 3DGS renderer, we implement the mapping from the state to the image by
\begin{equation}
    \hat\image= \R(\gauss^*_{\transf},\camera).
\end{equation}

\subsubsection{End-to-End Gradient Flow}
At this point, we have presented all of the components in our ContactGaussian-WM framework. 
This design allows for the end-to-end optimization of Eq.~\eqref{eq:prob}. 
Gradients from the image loss are backpropagated through the renderer $\R$ to the state, and subsequently propagated through the differentiable simulator $\dyn$ to optimize the physical parameters $\para$ and geometric parameters $\gauss_{\mathrm{geo}}$, which are collectively denoted as $\allpara$.

Mathematically, the gradient flow representing the backpropagation process through time can be written as follows:
\begin{equation}
    \frac{\dd \loss}{\dd \allpara} = \sum_{t \in \timeSet} \frac{\dd \loss}{\dd \hat\image} \cdot \frac{\partial \hat\image}{\partial \hat{\stat}_{t}} \frac{\dd \hat{\stat}_{t}}{\dd \allpara},
    \label{eq:gradient_flow}
\end{equation}
where $\frac{\dd \hat{\stat}_{t}^{(i)}}{\dd \allpara}$ is recursively computed via the differentiable simulator $\dyn$:
\begin{equation}
    \frac{\dd \hat{\stat}_{t}^{(i)}}{\dd \allpara} = \frac{\partial \dyn}{\partial \allpara} + \frac{\partial \dyn}{\partial \hat{\stat}_{t-1}^{(i)}} \frac{\dd \hat{\stat}_{t-1}^{(i)}}{\dd \allpara}.
    \label{eq:simulator_grad}
\end{equation}

Simultaneously optimizing the physical process and visually aligned images presents significant ambiguity, and images in motion are too blurry to adequately optimize our initial Gaussian; thus, as the formulation explicitly describes, the image loss does not update the geometry parameters in Gaussians directly through the rendering layer. 
Instead, these parameters are fine-tuned exclusively through the physical collision interactions propagated from the differentiable simulator.

\section{Results}
We evaluate our proposed framework in both simulation and real-world scenarios. 
We aim to answer the following  questions: 
\begin{enumerate}
    \item Can the proposed framework learn the underlying physical laws through sparse and contact-rich scene videos for more accurate predictions than existing methods?
    \item Can the method generalize to diverse real-world scenarios and support effective sim-to-real control transfer? 
    \item What is the utility of the proposed framework in downstream tasks?
\end{enumerate}

We leverage simulation to strictly benchmark the prediction accuracy of our approach.
Extending to the real world, we validate the model's generalization capability and sim-to-real control transfer across diverse real-world scenarios. 
Finally, to demonstrate the practical utility, we explore the potential of the proposed framework in supporting diverse downstream robotic applications.
Detailed setup and results can be found in the Appendix.

\subsection{Evaluations in simulation}
\textbf{Experimental Setup:} We initialized the experiment by converting the object's precise mesh into a 3DGS representation using our SG-GS method. 
Two interaction scenarios:
\begin{itemize}
    \item \Push: various objects pushed by a virtual hand with images, ground-truth trajectory, and end-effector velocity being recorded.
    \item \Fall: the complete free-fall and rebound process of various objects with images and ground-truth trajectories being recorded.
\end{itemize}
are designed within MuJoCo \cite{todorov2012mujoco} to generate trajectories along with their RGB renderings.
These two scenarios represent two distinct modes of motion: the former involves quasi-dynamic, continuous contact, while the latter features high-dynamic, discrete contact. 
To verify the framework’s generalization across various geometries and physical properties, we employ a diverse set of objects (e.g., Camera, Stanford Bunny, Piggy Bank, and Rubber Duck). 
The model is trained on a short interaction trajectory and evaluated across multiple unseen test sets. 
Specifically, for the falling scenario, the objects are released from various initial poses and positions; for the pushing scenario, the manipulator operates at different end-effector velocities. 
To assess long-horizon prediction stability, the test sequences are notably longer than those used during training, while maintaining a consistent sampling rate.

\textbf{Experimental metrics:}
Our evaluation focuses on the model's learning capability and predictive accuracy in an open-loop manner (i.e., only the true state of the first frame is provided). 
To assess learning performance, we track the \textbf{PSNR} of the rendered images during the training to monitor the model's ability to fit training data. 
Furthermore, we quantify predictive accuracy by averaging the translation and rotation errors between the ground-truth and predicted poses across the entire test set.

\textbf{Baseline methods:} We compare our proposed world model against various existing baselines, categorized as follows:

\textit{Data-driven world models.} We select DreamerV3~\cite{hafner2024masteringdiversedomainsworld} as a representative baseline for this category. 
As a general-purpose world model, DreamerV3 learns latent dynamics purely from visual observations but typically requires substantial data to converge. 
We train it using a dataset of 100 trajectories that are distinct from the test set, at the same time, we also follow its closed-loop training mode (conditioning on ground-truth states at every time step).

\textit{Physics-informed world models.} We select two representative approaches to cover the spectrum of physics-grounded optimization. 
The first is essentially a gradient-free system identification baseline. 
We integrate the standard MuJoCo simulator with our 3DGS renderer $\R$ and employ the gradient-free optimization strategy CEM from TwinTrack~\cite{yang2025twintrackbridgingvisioncontact} to minimize L1 image error, which we call it CEM+MuJoCo+$\R$ in the sequel. 
The second approach is PIN-WM~\cite{li2025pinwmlearningphysicsinformedworld}, representing differentiable physics-based methods, which constructs a physical-grounded world model with Linear Complementarity Problems (LCP) and utilizes implicit differentiation for efficient gradient-based optimization.

Notably, we provide realistic meshes as collision geometry for CEM+MuJoCo+$\R$ and PIN-WM, and utilize a high-fidelity physics simulator for accurate collision detection.
We also trained high-quality 2DGS objects for PIN-WM to prevent rendering artifacts from becoming a performance bottleneck. 
For a fair comparison, we used the official implementations with default hyperparameters for PIN-WM and DreamerV3. For CEM+MuJoCo+$\R$, we implemented 100 sets of parallel parameters using MJX \cite{todorov2012mujoco}, with the rest of the settings remaining the same as TwinTrack. 
All physics-grounded baselines are trained and evaluated on the same trajectory dataset as our proposed model, with each method trained until convergence to report its optimal performance.

\begin{figure}[t]
    \centering
    \begin{minipage}{0.48\linewidth}
        \centering
        \includegraphics[width=\linewidth]{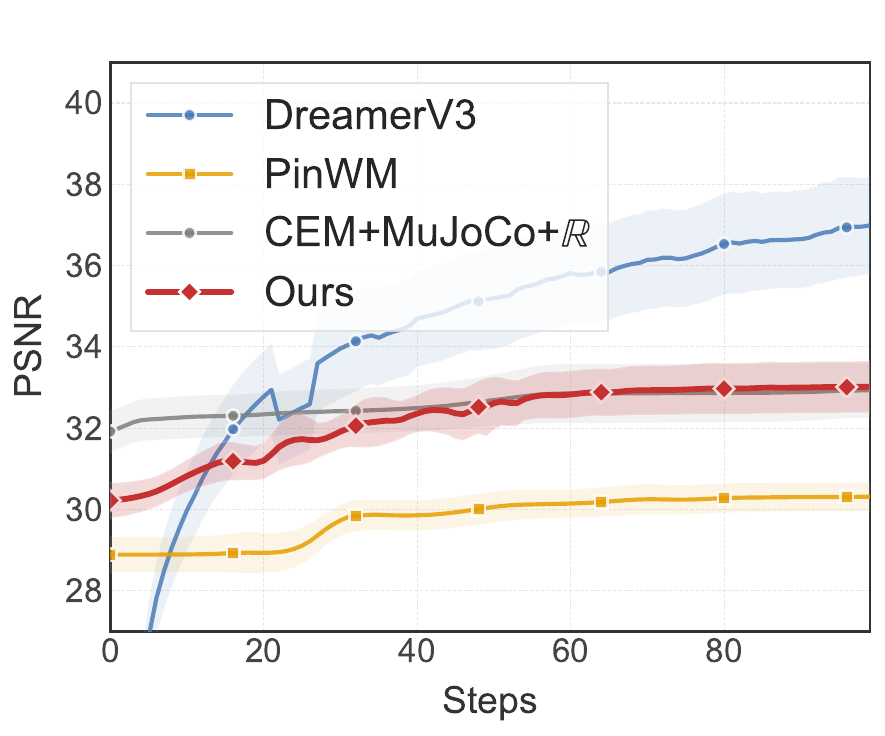} 
    \end{minipage}
    \hfill 
    \begin{minipage}{0.48\linewidth}
        \centering
        \includegraphics[width=\linewidth]{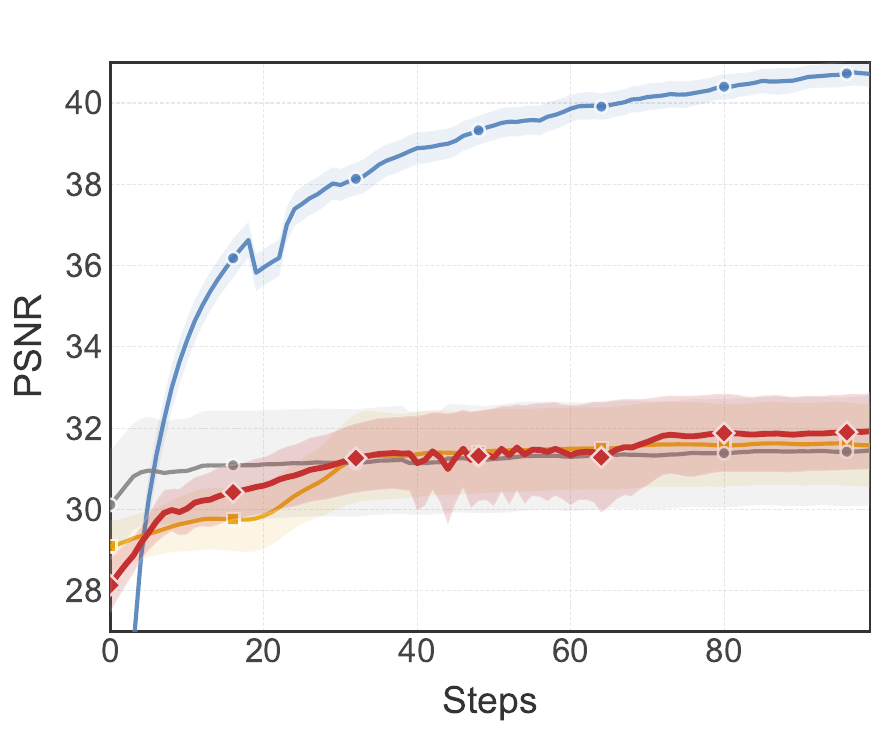} 
    \end{minipage}

    \vspace{0.2cm} 
        \caption{PSNR curves during training for \Fall~(left) and \Push~(right). DreamerV3 reports closed-loop one-step prediction, while the other methods report open-loop cumulative error.}
    \label{fig:trainloss}
\end{figure}

\textbf{Comparisons on training performance:}
Fig.~\ref{fig:trainloss} presents the learning performance curves. 
It can be observed that all four models converge in both scenarios. 
As DreamerV3 employs a closed-form training paradigm, it is reasonable that its PSNR significantly outperforms the other models. 
However, its convergence in the \Fall~scenario is slightly lower than in the \Push~scenario due to the higher visual complexity of the former, despite its lack of explicit consideration of contact dynamics.

For the three physics-based models, learning performance is comparable in the \Push~scenario. 
The reason is that this quasi-dynamic task provides smooth and continuous gradients, and the motion trajectories are predominantly governed by the friction coefficient. 
Such a constrained search space and stable constraints allow both the gradient-free CEM+MuJoCo+$\R$ and the gradient-based methods (PIN-WM and ours) to easily locate the optimal solutions.

In contrast, the \Fall~scenario presents a greater challenge. 
Collisions occur only within a few frames, providing limited useful information. 
Furthermore, discrete contacts lead to gradient discontinuities. 
PIN-WM, which already struggles with discrete and inaccurate gradients, fails to reach a satisfactory convergence under these conditions. 
For CEM+MuJoCo+$\R$, the parallel search allows them to fit the image loss very well.
In comparison, our model achieves a more stable and accurate gradient flow. 
By effectively utilizing sparse and discontinuous collision information, our framework can identify precise gradient directions with less computational resources, ultimately converging to a better value.

\textbf{Comparisons on test performance:}
To evaluate the predictive accuracy of the four models, we record the cumulative translation and orientation errors on the test set (Tables~\ref{table.Test results of the fall scenario} and ~\ref{table.Test results of the push scenario}), with qualitative results being presented in the Appendix. 
Based on the performance, we draw the following conclusions:

 1) The purely data-driven DreamerV3 suffers from a catastrophic decline in performance when generalizing to novel states and actions, failing to maintain essential scene consistency or physical plausibility during long-horizon forecasting. 
    
2) Although CEM+MuJoCo+$\R$ performs well in both \Push~and \Fall~scenarios, its generalization performance drops significantly in highly dynamic and discontinuous fall scenarios, indicating that the model is unable to capture the underlying physical laws in these scenarios when the scene changes greatly, and instead overfits the training loss.
 
3) Frameworks like PIN-WM, which rely on implicit function differentiation and non-differentiable collision detection, are ill-suited for high-dynamic scenarios. 
The inherent inaccuracy and lack of smoothness in their gradient estimation make it difficult to derive reliable update directions from sparse and highly dynamic signals. 
Consequently, such methods may function in quasi-dynamic scenarios with continuous contacts (e.g., pushing).

4) In contrast, our proposed ContactGaussian-WM framework enables end-to-end differentiable learning. This allows the model to accurately capture underlying physical laws from images, even under challenging conditions such as inaccurate geometry, imperfect dynamics, sparse contact information, and complex collision interactions.
Consequently, our framework achieves stable generalization performance. 
Furthermore, it also performs well in promoting learning and generalization in various scenarios, which further demonstrates the robustness and universality of our learning paradigm.

\begin{table*}[t]
    \centering
    \caption{Test results of the \Fall~scenario. 
    In the \Fall~scenario, our method consistently outperforms the baseline across all metrics and objects.}
    \label{table.Test results of the fall scenario}
    \begin{threeparttable}
        \setlength{\tabcolsep}{4pt}
        \begin{tabular}{lcccccccc}
            \toprule
            \multirow{2}{*}{Object} & 
            \multicolumn{2}{c}{Ours} &
            \multicolumn{2}{c}{PIN-WM} & 
            \multicolumn{2}{c}{CEM+MuJoCo+$\R$} &
            \multicolumn{2}{c}{DreamerV3} 
            \\
            \cmidrule(lr){2-3}
            \cmidrule(lr){4-5} 
            \cmidrule(lr){6-7}
            \cmidrule(lr){8-9}
             &
             Trans. [m] $\downarrow$ &
             Orient. [rad] $\downarrow$ &  
             Trans. [m] $\downarrow$ &
             Orient. [rad] $\downarrow$ &  
             Trans. [m] $\downarrow$ &
             Orient. [rad] $\downarrow$ &  
             Trans. [m] $\downarrow$&
             Orient. [rad] $\downarrow$ 
            \\ 
            \midrule
            Camera
             & \textbf{0.0116} & \textbf{0.0665} 
             & 0.0151& 0.305 
             & 0.041 & 0.197 
             & 0.101& 1.489\\ \addlinespace
            Stanford Bunny
             & \textbf{0.0048} & \textbf{0.0114} 
             & 0.0351& 0.340 
             & 0.0343& 0.0984& 0.0693& 1.91\\ \addlinespace
            Piggy Bank
             & \textbf{0.0139} & \textbf{0.0522}& 0.0167& 0.186& 0.0515& 0.164 
             & 0.056 & 1.53\\ \addlinespace
            Binoculars
             & \textbf{0.0141} & \textbf{0.0934} 
             & 0.0199& 0.347& 0.0374& 0.464& 0.112& 2.29 
             \\ \addlinespace
            Rubber Duck
             & \textbf{0.0498}& \textbf{0.0771}& 0.107 & 0.122& 0.104& 0.147 
             & 0.123& 1.85\\
            \bottomrule
        \end{tabular}
    \end{threeparttable}
    \vspace{-5pt}
\end{table*}

\begin{table*}[t]
    \centering
    \caption{Test results of the \Push~scenario. 
    In this scenario, our method achieves comparable results to baselines.}
    \label{table.Test results of the push scenario} 
    \begin{threeparttable}
        \setlength{\tabcolsep}{4pt}
        \begin{tabular}{lcccccccc}
            \toprule
            \multirow{2}{*}{Object} & 
            \multicolumn{2}{c}{Ours} &
            \multicolumn{2}{c}{PIN-WM} & 
            \multicolumn{2}{c}{CEM+MuJoCo+$\R$} &
            \multicolumn{2}{c}{DreamerV3} 
            \\
            \cmidrule(lr){2-3}
            \cmidrule(lr){4-5} 
            \cmidrule(lr){6-7}
            \cmidrule(lr){8-9}
             &
             Trans. [m] $\downarrow$ &
             Orient. [rad] $\downarrow$ &  
             Trans. [m] $\downarrow$ &
             Orient. [rad] $\downarrow$ &  
             Trans. [m] $\downarrow$ &
             Orient. [rad] $\downarrow$ &  
             Trans. [m] $\downarrow$&
             Orient. [rad] $\downarrow$ 
            \\ 
            \midrule
            Camera
             & 0.0043 & \textbf{0.0114} 
             & \textbf{0.0027} & 0.0260 
             & 0.0168 & 0.0839 
             & 0.0790& 0.332\\ \addlinespace
            Stanford Bunny
             & 0.0044& 0.0196& \textbf{0.0041} & \textbf{0.0127}
             & 0.0059 & 0.0032& 0.1085& 0.0523\\ \addlinespace
            Piggy Bank
             & 0.0074& 0.0394& \textbf{0.0037} & 0.0225 
             & 0.0058 & \textbf{0.0189} 
             & 0.111& 0.0825\\ \addlinespace
            Binoculars
             & \textbf{0.0055}& \textbf{0.02}& 0.0061 & 0.0225 
             & 0.0059 & 0.2300 
             & 0.0901& 0.324\\ \addlinespace
            Rubber Duck
             & 0.0218 & 0.0367 
             & 0.0164 & 0.0730 
             & \textbf{0.0128} & \textbf{0.0136} 
             & 0.0898& 0.0354\\
            \bottomrule
        \end{tabular}
    \end{threeparttable}
    \vspace{-5pt}
\end{table*}

\subsection{Evaluations in real world}
While the quantitative and qualitative performance of our model against various baselines has been established in simulation, in this part, we shift our focus towards validating practical feasibility in complex, diverse environments and the effectiveness of sim-to-real transferability.

\begin{figure}[t] 
    \centering
    \includegraphics[width=1.0\linewidth]{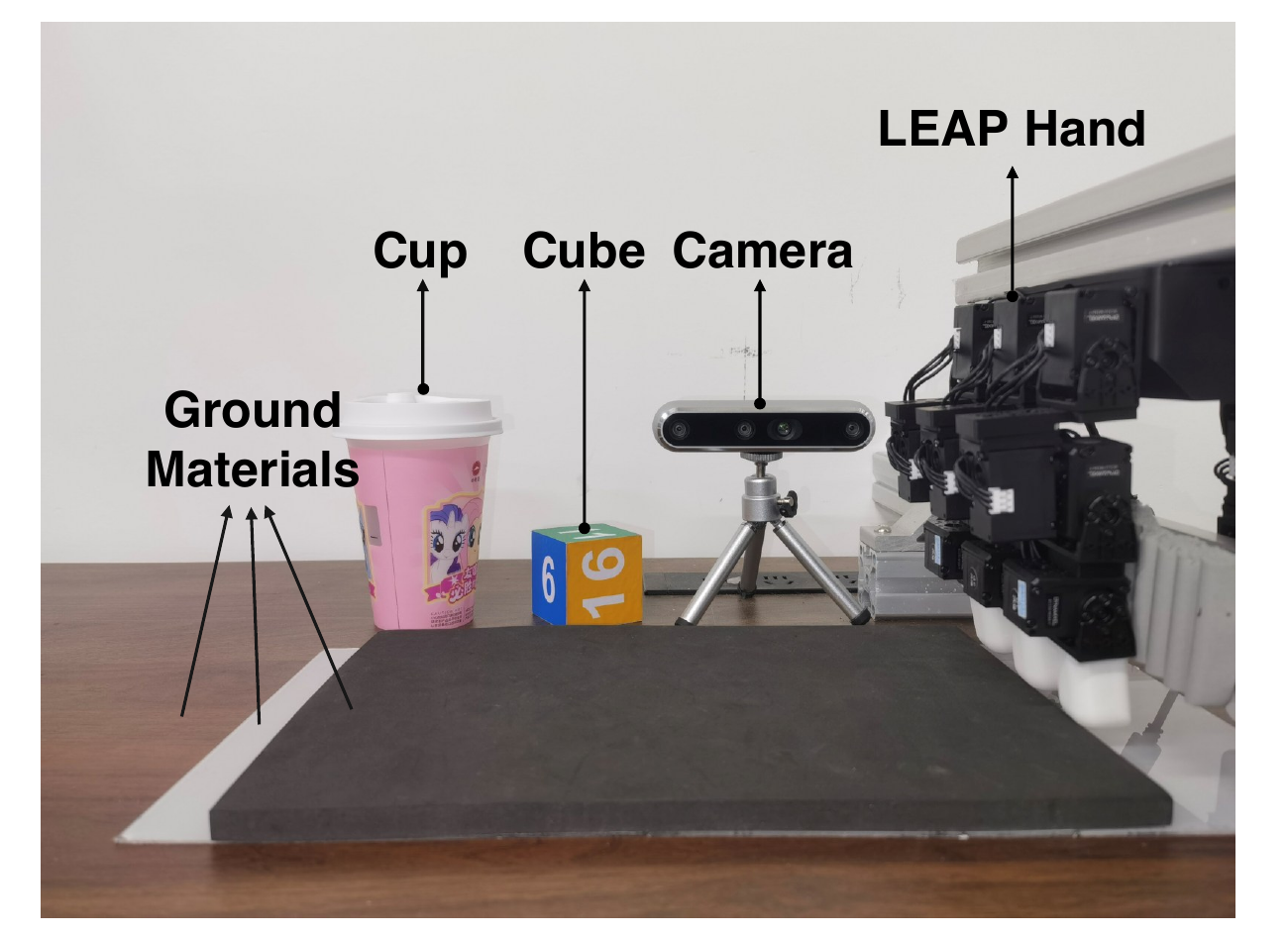} 
    \caption{Our real-world experiment setup}
    \label{fig:handwareset}
\end{figure}

\textbf{Experimental Setup:} 
We design two dynamic scenarios: free-fall and robotic manipulation. 
In the latter, we use the LEAP Hand~\cite{shaw2023leaphandlowcostefficient} which exerts high-impact forces to induce complex behaviors like sliding and flipping, as seen in Fig.~\ref{fig:handwareset}. 
Following \cite{pang2021convex}, we model the hand as a spring system under impedance control~\cite{hogan1984impedance} to simplify contact dynamics. 
Additionally, we vary ground materials (rubber, PVC, and wood) to test robustness against diverse physical properties (e.g., stiffness, friction). 
We initialize 3DGS representations using SG-GS from multi-view images captured by hand phone.
We capture motion videos for each trial using a RealSense D456 RGBD camera, splitting them into training and test sets (1:4 ratio). 
Unlike methods relying on external state tracking, we treat the initial state as a learnable parameter. 
To decouple it from geometry and physics optimization, we pre-estimate the initial state using the first few frames. 
Finally, we implement an open-loop sim-to-real transfer strategy, executing trajectories which are planned based on the model on the LEAP hand to validate whether it supports valid control transfer.

\textbf{Experimental metrics:} Due to the lack of reliable real-world state information, we use the cumulative peak signal-to-noise ratio (PSNR) of long-horizon predictions as the primary metric. 
The high visual fidelity of long-horizon open-loop predictions strongly indicates that the model has correctly captured the underlying physical laws and achieved accurate dynamic predictions. 
For prediction experiments on the test set, we calculate the average cumulative PSNR of open-loop predictions. 
For each scene, we conduct multiple experiments with different materials and then calculate the average. 
Similarly, for open-loop trajectory transfer experiments, we conduct multiple experiments with different materials for each object and calculate the average PSNR between the actual and planned images.

\textbf{Baseline method:}
To highlight the impact of parameter learning on model prediction and sim-to-real transfer, we compare it with ContactGaussian-WM without parameter learning, while keeping all other variables constant.

\begin{figure}[t] 
    \centering
    \includegraphics[width=1.0\linewidth]{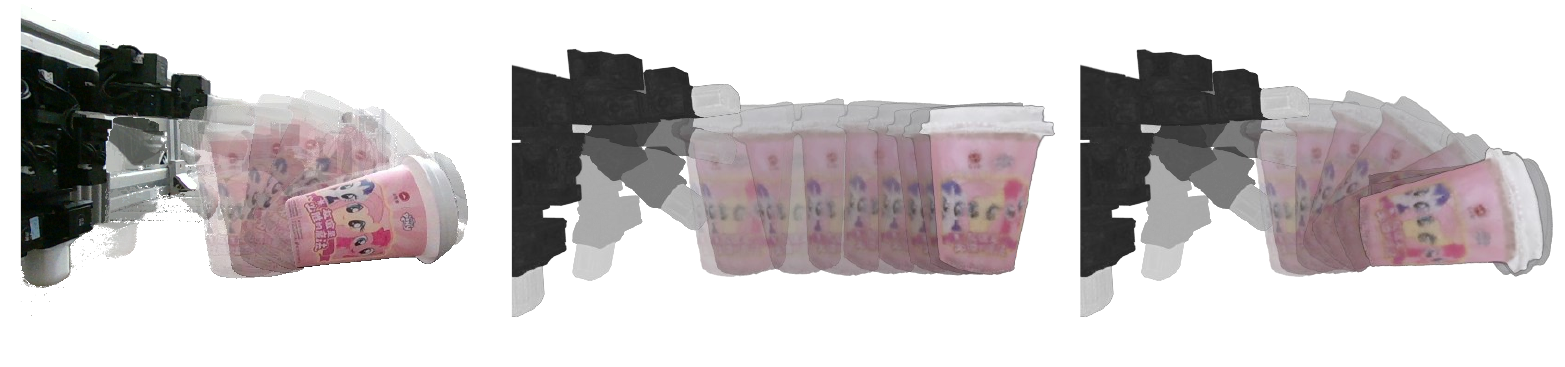} 
    \caption{Visualization of long-horizon predictions in real-world experiments. 
    Our method (right) outperforms the one without optimization (middle), as compared with the ground-truth observation (left).}
    \label{fig:future_pred}
\end{figure}

\textbf{Results:}
In Tables~\ref{table.Table of Long-Range Prediction Results from Real Experiments} and \ref{table.OpenLoopTransfer_PSNR_Real}, we report the performance of the two methods on the test set and in actual open-loop trajectory transfer, respectively. 
Fig.~\ref{fig:future_pred} visualizes the actual performance of the two methods in the open-loop controlled trajectory transfer experiment.
It can be found that our world model parameter learning improves the prediction fidelity of the world model while narrowing the sim-to-real gap.

\begin{table}[t]
    \centering
    \caption{Long-horizon prediction results from real-world experiments. Values represent PSNR [dB].}
    \label{table.Table of Long-Range Prediction Results from Real Experiments}
    \begin{threeparttable}
        \setlength{\tabcolsep}{5pt} 
        \begin{tabular}{lcccc}
            \toprule
            Methods & Cube\_fall & Cube\_LeapHand & Cup\_fall & Cup\_LeapHand \\
            \midrule
            Ours & \textbf{27.144} & \textbf{27.180} & \textbf{20.90} & \textbf{19.41} \\ \addlinespace
            No\_opt & 25.01 & 25.35 & 18.19 & 17.046 \\ \addlinespace
            \bottomrule
        \end{tabular}
    \end{threeparttable}
    \vspace{-5pt}
\end{table}

\begin{table}[t]
    \centering
    \caption{Open-Loop trajectory transfer results from real-world experiments. Values represent PSNR [dB].}
    \label{table.OpenLoopTransfer_PSNR_Real}
    \begin{threeparttable}
        \setlength{\tabcolsep}{5pt} 
        \begin{tabular}{lcccc}
            \toprule
            Methods & Cube\_fall & Cube\_LeapHand & Cup\_fall & Cup\_LeapHand \\
            \midrule
            Ours   & \textbf{25.92} & \textbf{26.03} & \textbf{19.62} & \textbf{18.12} \\ \addlinespace
            No\_opt & 22.28 & 22.41 & 15.78 & 14.46 \\ \addlinespace
            \bottomrule
        \end{tabular}
    \end{threeparttable}
    \vspace{-5pt}
\end{table}

\subsection{Applications}
\textbf{Simulation data synthesis:} 
A pivotal application of world models lies in simulation data synthesis. 
Extensive evaluations across both simulated and real-world settings have demonstrated our model's robust capability in understanding complex physical interactions and forecasting dynamics, particularly in contact-rich dynamic and high-dynamic scenarios. 
Consequently, the trained ContactGaussian-WM serves as a powerful engine for generating physically accurate and visually plausible simulation data, enabling the construction of high-fidelity 4D environments. 
Furthermore, our framework achieves an image generation rate of approximately 40Hz on an RTX 4090. 
This computational efficiency highlights the potential for real-time sim-to-real interaction.

\begin{figure}[t] 
    \centering
    \includegraphics[width=1.0\linewidth]{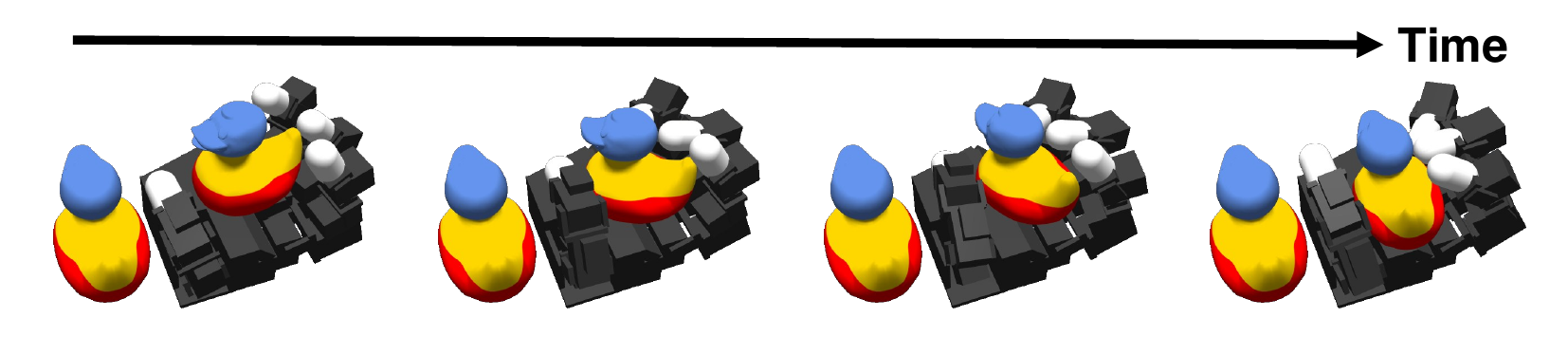} 
    \caption{Time-series visualization of applying MPC on the LEAP Hand to reorient the rubber duck in MuJoCo.}
    \label{fig:mpc_duck}
\end{figure}

\textbf{Real-time MPC control:}
The differentiability and explicit nature of the ContactGaussian-WM framework enable seamless integration with existing MPC solvers.
In this experiment, we first train a world model based on target objects using MuJoCo synthetic interactive data, and then implement a real-time MPC framework. 
As shown in the Fig.~\ref{fig:mpc_duck}, our framework enables LEAP Hand to successfully perform in-hand redirection tasks in MuJoCo, accurately planning and executing the control sequences required to manipulate objects.
\section{Conclusion and future works}
We have proposed ContactGaussian-WM, an end-to-end differentiable framework for reconstructing world models from sparse videos. 
Compared to other methods, our world model is better suited for learning from highly dynamic, contact-rich videos, and we have demonstrated the effectiveness of our framework in a variety of scenarios. 
Furthermore, we have showcased its use for simulation data synthesis and integration with real-time MPC.

\textbf{Limitation and future work:} 
We identified several promising directions for future work. 
First, the proposed unified scene representation can, to some extent, reduce rendering quality. 
Second, Gaussian-sphere-based collision detection may fail to provide accurate penetration distances and contact normals under deep interpenetration. 
Finally, our current framework is limited to rigid-body systems and does not include deformable objects. 
Future work will explore more visually faithful and geometrically accurate unified representations, more accurate differentiable collision detection, and more general world models that extend beyond rigid bodies.

\newpage
\bibliographystyle{IEEEtran}    

\bibliography{main-arxiv}
\clearpage
\appendix
\subsection{Additional Details for Scene Initialization with Sphere Geometry Gaussian Splatting (SG-GS)}
\begin{figure*}[t] 
    \centering
    \includegraphics[width=1.0\textwidth]{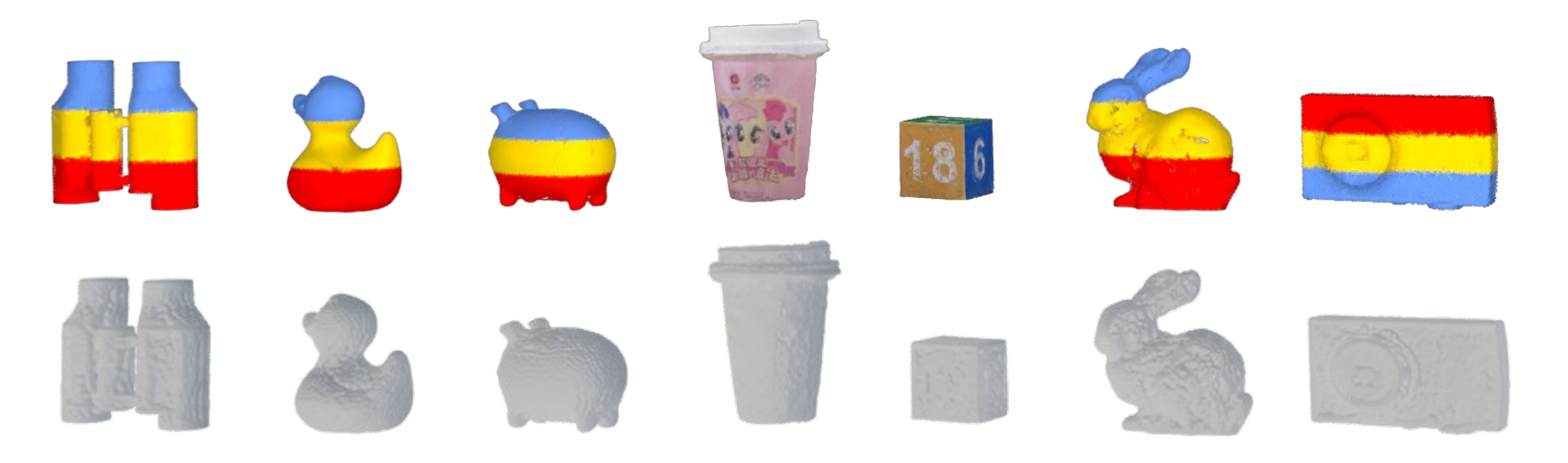} 
    \caption{Visualization of Unified Representation. (Top) Gaussian Splats generated by SG-GS. (Bottom) Extracted OBJ meshes derived from the implicit SDF via our Soft Distance Detection and Marching Cubes \cite{lorensen1998marching}}
    \label{fig:sg-gs_performance}
\end{figure*}

For synthetic objects and the components of LEAP Hand, we use Blender to render OBJ files from 72 uniform spherical views, acquiring images, masks, camera parameters, and mesh-based dense point clouds. 
For real-world objects, we estimate camera poses using COLMAP \cite{schonberger2016structure} on the multi-view images and masks. 
To accelerate 3DGS geometric prior optimization, we replace the standard sparse point cloud with a dense one initialized via Visual Hull \cite{273735}. 
Fig.\ref{fig:sg-gs_performance} illustrates the unified representations produced by SG-GS.

\subsection{Additional Details for Differentiable Collision Detection}
\begin{figure}[H] 
    \centering
    \includegraphics[width=1.0\linewidth]{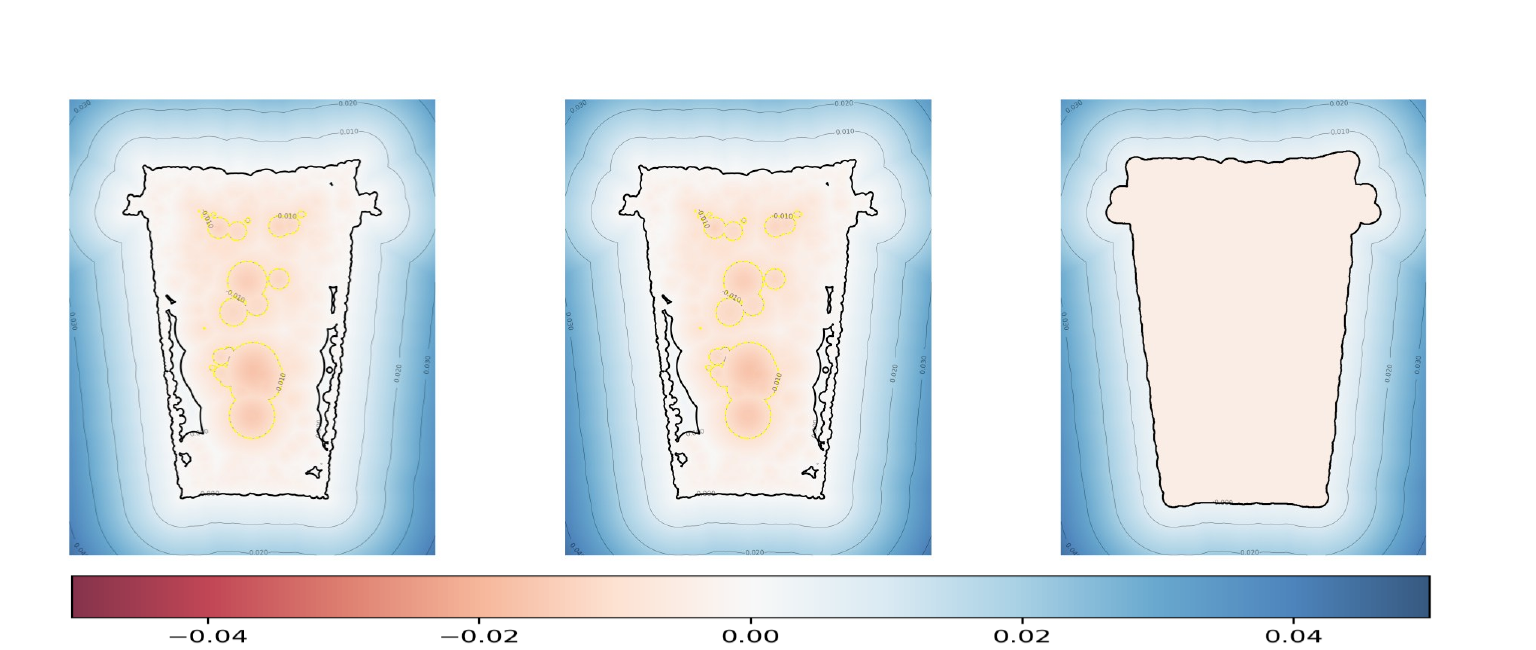} 
    \caption{Visualization of penetration distance processing for the cup object. 
    We visualized SDF slices computed using the hard minimum (left), the LogSumExp (middle), and the LogSumExp with penetration handling (right).}
    \label{fig:collision_detection}
\end{figure}

As mentioned in the main paper, we employ LogSumExp for sphere collision detection.
However, this approach fails to yield accurate penetration depths. 
Therefore, we use a Sigmoid function to enforce a fixed distance when penetration occurs, thereby ensuring stable detection. Fig.\ref{fig:collision_detection} illustrates the impact of this modification.

We observe that the LogSumExp and hard-min results are almost identical. 
However, when penetration occurs inside the object, the LogSumExp formulation can sometimes return an incorrectly small penetration value, which can further exacerbate penetration. 
After adding penetration handling, the predicted penetration distance inside the object is effectively bounded, reducing severe penetrations.
Since we do not need to optimize the scene itself, we further simplify the collision-checking environment. 
For example, we model the ground as a plane and downsample the robotic hand with a complex structure to a smaller set of query points.
These simplifications reduce both model and computational complexity, leading to more stable optimization.

\subsection{Additional Details for Differentiable Complementarity-free Contact Dynamics}
For the Jacobian matrix and collision distance obtained($\jac$, ${\Dist}$) from collision detection, we follow the same approach as in \cite{jin2024complementarity}. Specifically, we project $\jac$ onto the normal vector $\nvec$ and $n_{d}$ tangent vectors $\dvec$, and then stack them to form $\tilde\jac(\mu)$, and stack
the collision distance ${\Dist}$ into $\tilde{\Dist}$ accordingly:
\begin{equation*}\label{equ.def_mats}
\small
\begin{aligned}
\tilde\jac(\mu):=
\begin{bmatrix}
    \jac_{1}^\nvec{-}\mu_1\jac_{1,1}^\dvec
    \\
    \cdots
\\
\jac_{1}^\nvec{-}\mu_{1}\jac_{1,n_d}^\dvec\\
\vdots\\
\jac_{n_c}^\nvec{-}\mu_{n_c}\jac_{n_c,1}^\dvec
    \\
    \cdots
\\
\jac_{n_c}^\nvec{-}\mu_{n_c}\jac_{n_c,n_d}^\dvec
\end{bmatrix},
\,\,\,\,
\tilde{\Dist}:=\begin{bmatrix}
    \Dist_{1}
    \\
    \cdots
\\
\Dist_{1}\\
\vdots\\
\Dist_{n_c}
    \\
    \cdots
\\
\Dist_{n_c}
\end{bmatrix},
\end{aligned}
\end{equation*}
where $\jac_{n_c}^\nvec$ denotes the component of the $\jac$ at the $n_c$ contact point along the normal direction,and $\jac_{n_c,1}^\dvec$ denotes the component of the $\jac$ at the $n_c$ contact point along the $n_d$ tangential direction.

\subsection{Additional Details for Simulation Experiment}
\subsubsection{Datasets} 
\begin{figure}[t] 
    \centering
    \includegraphics[width=1.0\linewidth]{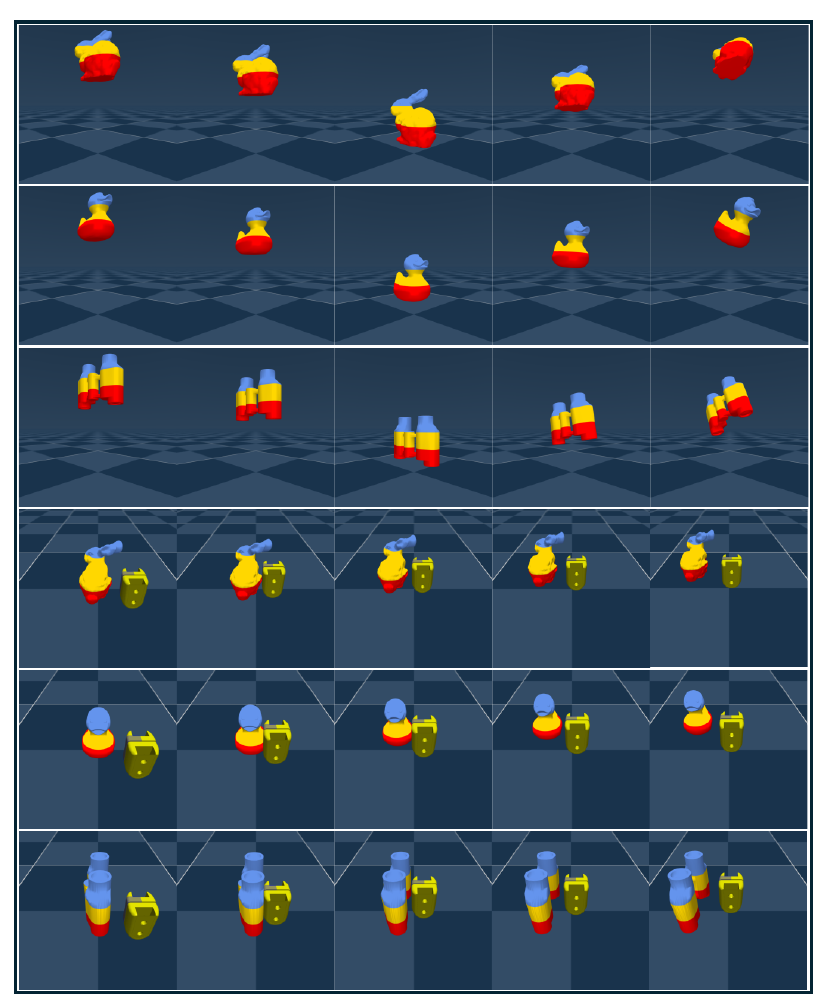} 
    \caption{Visualization of some simulated object data}
    \label{fig:sim_datasets}
\end{figure}

We used a fixed camera pose to record the complete motion trajectory in MuJoCo, including the initial state of the object, the camera's intrinsic and extrinsic parameters, and the motion trajectory of the end effector. Fig.\ref{fig:sim_datasets} visualizes the simulation dataset.

\subsubsection{Implementation Details of Baselines}
Due to the page limit in the main paper, here we provide additional implementation details for baselines.
To ensure a fair comparison, we use the same training dataset for ContactGaussian-WM, PIN-WM\cite{li2025pinwmlearningphysicsinformedworld}, and CEM+MuJoCo+$\R$, consisting of 15 frames. 
For DreamerV3\cite{hafner2024masteringdiversedomainsworld}, we randomize the initial states from this training set to generate 100 training sequences, and extend each sequence to 25 frames. 
The test set used for reporting metrics is kept identical across all methods.
During training, for methods that require initialization of physical parameters (ContactGaussian-WM, PIN-WM, and CEM+MuJoCo+$\R$), we apply the same randomization range to shared physical parameters to obtain an initial parameter set; for CEM+MuJoCo+$\R$, we sample 100 initial parameter sets within this range.
In addition, we provide the exact formulas for the quantitative evaluation metrics, average translation error, average rotation error, and average PSNR, which are averaged over $N$ sequences and frames $T$.
\begin{equation*}
\begin{aligned}
    E_{trans} &= \frac{1}{NT} \sum_{i=1}^{N} \sum_{t=1}^{T} \|\pos_{i,t} - \hat\pos_{i,t}\|_2, \\
    E_{rot} &= \frac{1}{NT} \sum_{i=1}^{N} \sum_{t=1}^{T} 2 \arccos(|\quat_{i,t} \cdot \hat\quat_{i,t}|), \\
    E_{psnr} &= \frac{1}{NT} \sum_{i=1}^{N} \sum_{t=1}^{T} \left[-10\log_{10}\!\left(\frac{1}{M}\sum_{p=1}^{M}(I_{i,t}^{(p)}-\hat I_{i,t}^{(p)})^2\right)\right].
\end{aligned}
\end{equation*}
\subsubsection{Supplement to Qualitative Results}
\begin{figure*}[t] 
    \centering
    \includegraphics[width=1.0\textwidth]{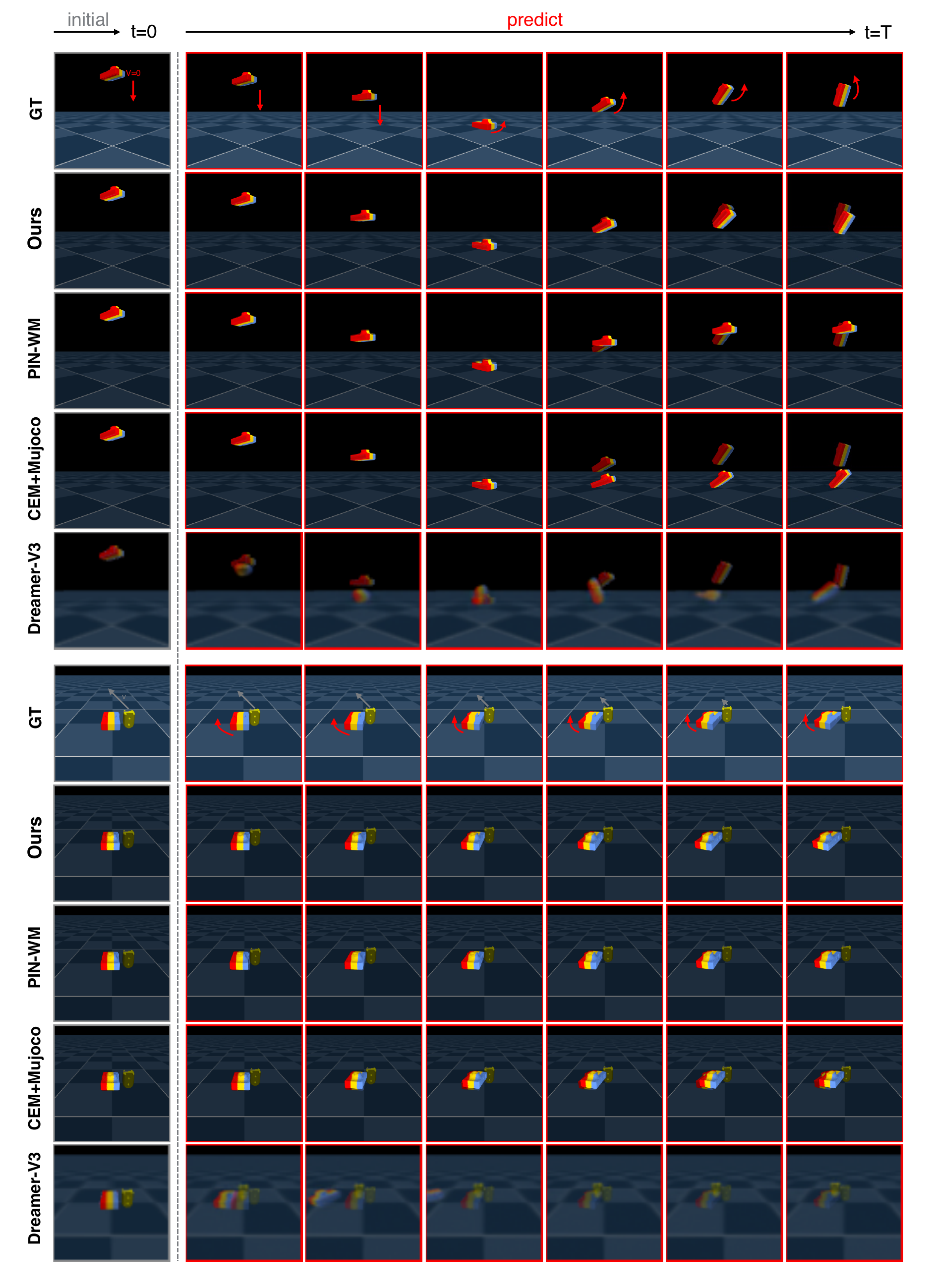} 
    \caption{Visualization of test results on the camera object for all methods in the simulation experiments.}
    \label{fig:sim_predict_camera}
\end{figure*}

We qualitatively demonstrate the performance of each baseline on the camera object in Fig.\ref{fig:sim_predict_camera}.

The performance of each method in two simulation scenarios further reflects the superior performance of our approach on \Push~tasks, while significantly outperforming other baseline methods on \Fall~tasks.

\subsection{Additional Details for Real World Experiments}
In real-world experiments, we first estimate the camera pose in the world coordinate system using ArUco markers \cite{GARRIDOJURADO20142280}.
For experiments involving the LEAP Hand, we adopt an inverted-Gaussian approach: we align rendered images of key LEAP Hand parts at a specified configuration with real images to recover the LEAP Hand pose in the world frame. 
Then, we use position control to implement velocity-like control for the LEAP Hand.
Specifically, given a target joint position, we limit the maximum joint velocity; once the LEAP Hand reaches this maximum speed, we record the current joint state as the initial joint configuration and start recording frames, ensuring that no contact with the object has occurred at that moment.

Moreover, as described in the main paper, without any additional state estimation module, we use our differentiable framework to estimate the initial state and initial velocity. 
Concretely, we freeze all irrelevant parameters and optimize the initial pose by aligning the target object in the first frame. 
We then fix the optimized pose and estimate the initial velocity by aligning the first three frames (where we ensure no contact occurs). 
This process of initial velocity estimation is only used in the free-fall scene.
As a result, our training set includes RGB images, camera parameters, initial pose, and initial velocity. 
For LEAP Hand experiments, it additionally contains the initial and target hand states, as well as the commanded joint motion speed.

\subsection{More Results for Real-time MPC control}
\begin{figure}[H] 
    \centering
    \includegraphics[width=1.0\linewidth]{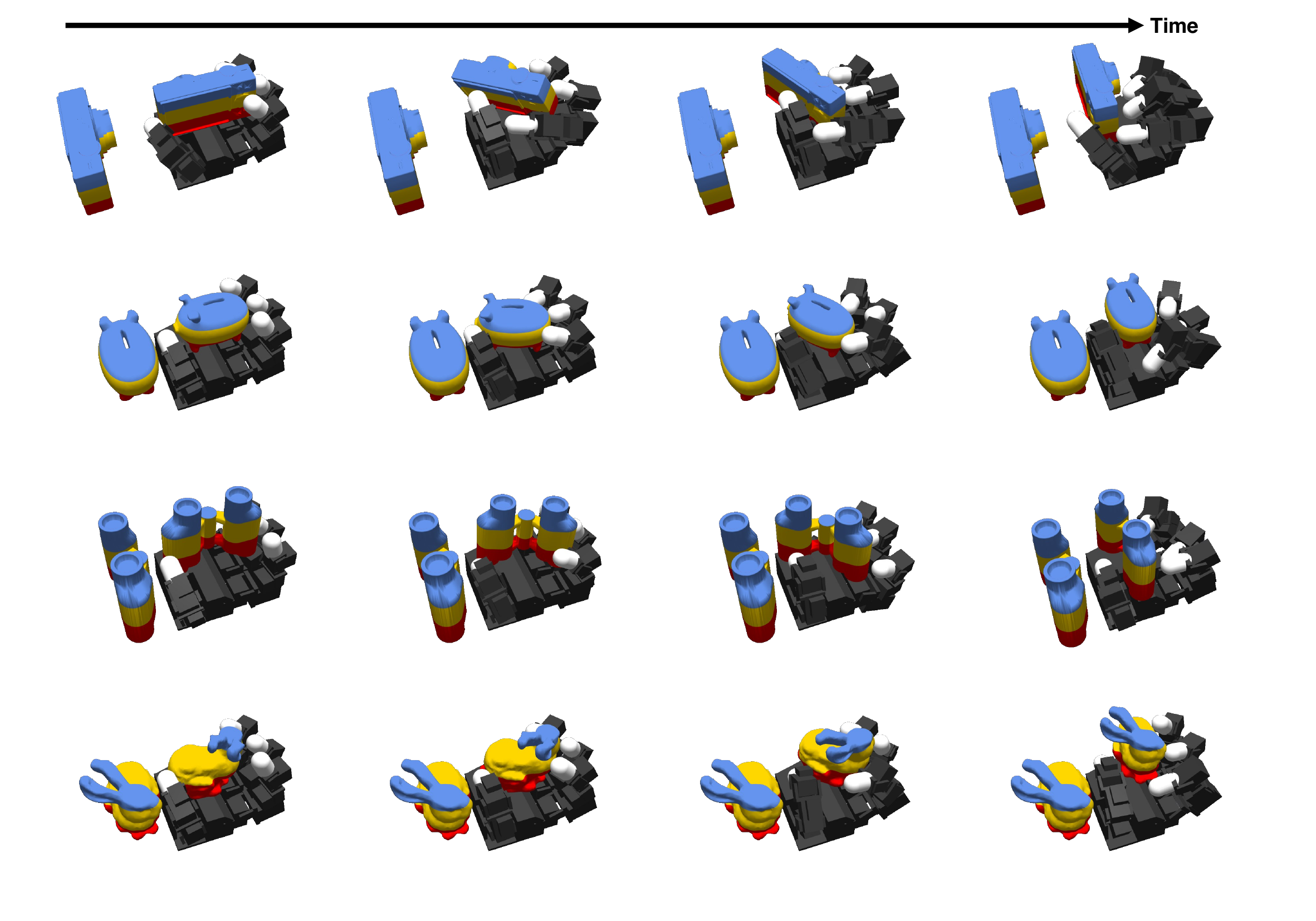} 
    \caption{More results of real-time MPC redirection experiment}
    \label{fig:mpc_appendix}
\end{figure}

We perform real-time MPC control in MuJoCo on different objects to achieve palm redirection for the LEAP Hand, as shown in Fig.\ref{fig:mpc_appendix}.

\end{document}